%% file: main_fix.tex
\documentclass[10pt,twocolumn,letterpaper]{article}
\usepackage{float}
\usepackage{multirow}

\usepackage{xcolor}
\usepackage{colortbl}
\usepackage{array}
\usepackage{booktabs}
\usepackage{tabularx}
\usepackage[T1]{fontenc}
\usepackage{graphicx}
\usepackage{booktabs}
\usepackage{amsmath,amssymb,mathtools}
\usepackage{microtype}
\usepackage{array}
\usepackage{xspace}

\usepackage{tikz}
\usepackage{pgfplots}
\usepackage{pgfplotstable}
\usepackage{subcaption}
\usepackage{cuted}
\usepackage{float}
\usepackage{placeins}

\definecolor{AvgTint}{HTML}{F0F5FE}

\definecolor{cvprblue}{rgb}{0.21,0.49,0.74}
\usepgfplotslibrary{groupplots}
\pgfplotsset{compat=1.18}

\definecolor{draftgray}{rgb}{0.95,0.96,0.98}

\definecolor{ablBlue}{HTML}{4D73BC}
\definecolor{ablOrange}{HTML}{DE8B38}
\definecolor{ablPurple}{HTML}{8A6AB5}
\definecolor{ablGray}{HTML}{8B919B}

\pgfplotsset{
    ablAxis/.style={
        width=0.97\linewidth,
        height=0.67\linewidth,
        axis line style={black!55},
        tick style={black!45},
        tick align=outside,
        grid=major,
        major grid style={black!9},
        minor tick num=0,
        tick label style={font=\scriptsize},
        label style={font=\small},
        legend style={
            font=\scriptsize,
            draw=none,
            fill=white,
            cells={anchor=west},
            inner sep=2pt
        },
        every axis plot/.append style={line width=1pt},
        scaled ticks=false
    }
}

\ifdefined\TrustTouchReview
    \usepackage[review]{cvpr}
\else
    \usepackage[pagenumbers]{cvpr}
\fi

\usepackage[pagebackref,breaklinks,colorlinks,allcolors=cvprblue]{hyperref}

\newcolumntype{L}[1]{
    >{\raggedright\arraybackslash}p{#1}
}

\title{
TipDex-VTLA: Learning to Refine Actions with Visual Reliability and Tactile Feedback for Dexterous Manipulation
}

\author{
Xudi Ge\textsuperscript{1,\#} \quad
Yepeng Liu\textsuperscript{2,\#} \quad
Haifeng Liu\textsuperscript{2,\#} \quad
Enze Shi\textsuperscript{1} \quad
Liangqi Zhang\textsuperscript{2} \quad
Yongchao Xu\textsuperscript{1,*}
\\[0.5em]
{\small
\textsuperscript{1}School of Computer Science, Wuhan University, Wuhan, China
}
\\[0.2em]
{\small
\textsuperscript{2}Wuhan Huaweike Intelligent Technology Co., Ltd.,
Wuhan, China
}
}

\def\paperID{DRAFT}
\def\confName{CVPR}
\def\confYear{2027}

\begin{document}

\maketitle

\begin{NoHyper}
\renewcommand{\thefootnote}{}
\footnotetext{%
  \textsuperscript{\#}Equal contribution.
  \quad
  \textsuperscript{*}Corresponding author.
  \par
  Xudi Ge's work was done during an internship at
  Wuhan Huaweike Intelligent Technology Co., Ltd.
}
\end{NoHyper}
% --------------------------------------------------
% Motivation figure
% --------------------------------------------------

\begin{strip}
    \centering
    \includegraphics[
        width=0.9\textwidth
    ]{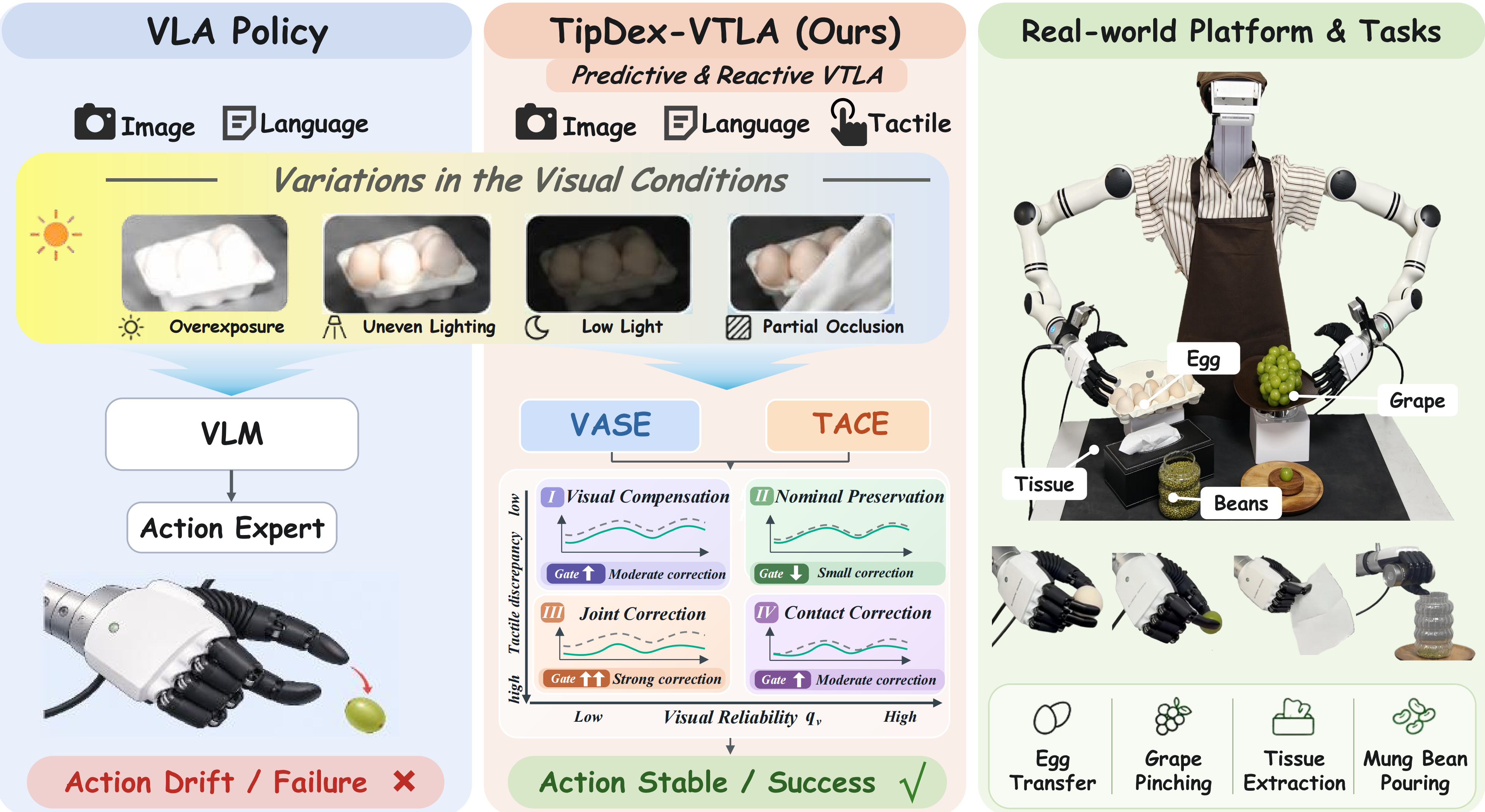}
    \captionsetup{hypcap=false}
    \captionof{figure}{
        \textbf{Action refinement with visual reliability and tactile feedback.}
  \textbf{Left:} Visual disturbances can cause VLA action drift
despite an unchanged physical scene.
\textbf{Center:} Visual reliability and tactile forecast agreement
motivate four complementary regimes for execution-time refinement
in TipDex-VTLA.
\textbf{Right:} Our robotic platform and four dexterous
manipulation tasks with TipDex-VTLA.}
    \label{fig:motivation}
\end{strip}

% --------------------------------------------------
% Abstract
% --------------------------------------------------
\begin{abstract}
Changes in illumination can induce action drift in dexterous manipulation even when the physical state remains unchanged. Although existing vision--tactile--language--action (VTLA)
policies incorporate tactile feedback, when and
how touch refines actions under changing visual
conditions remains challenging.
We introduce \textbf{TipDex-VTLA}, a VTLA policy that combines estimated 
visual action reliability and tactile forecast verification
to regulate execution-time refinement. Specifically,
Visual Action Sensitivity Estimation (VASE) learns visual
reliability from a frozen reference policy's action sensitivity
to illumination perturbations.
Tactile Anticipation and Correction during Execution (TACE)
compares predicted and measured touch at matched times and
combines the resulting discrepancy with visual reliability
to gate bounded feature corrections for unexecuted actions. This effectively helps to refine actions under challenging unseen visual conditions. We evaluate TipDex-VTLA on four real-world dexterous manipulation tasks and the UniVTAC simulation benchmark. TipDex-VTLA trained on data collected under normal visual conditions, achieves average
full-task success rates of 87.5\% (\textit{resp.} 73.8\%) under normal (\textit{resp.} challenging unseen) lighting in real-world tasks, significantly outperforming some state-of-the-art methods. We will release the collected data and source code. 
\end{abstract}

\input{sec/introduction}

\input{sec/relatedwork}

\input{sec/method}

\input{sec/experiments_fix}

\input{sec/conclusion}
% --------------------------------------------------
% References
% --------------------------------------------------
\clearpage
{\small
\bibliographystyle{ieeenat_fullname}
\bibliography{references}
}

\end{document}

%% file: sec/introduction.tex
\section{Introduction}

Vision--Language--Action (VLA) policies generate manipulation actions
from visual observations and language instructions~\cite{openvla,octo,pi0,pi05}.
In contact-rich dexterous manipulation, however, the hand often
occludes the interaction region, making the onset and evolution
of local contact difficult to observe. Besides,
visual guidance is also sensitive to illumination: lighting changes
can alter policy predictions even when the physical scene remains
unchanged~\cite{liberoplus,robustvla,stablevla,egr}.
Their impact depends on action-relevant cues: a dim image may
preserve grasp geometry, whereas a localized reflection may
obscure the intended contact region.

Vision--Tactile--Language--Action (VTLA) policies complement visual
observations with tactile feedback during action generation and
execution~\cite{atvla,trex,unitacvla}.
For example, fingertip pressure is expected to increase when
grasping an egg. An increase beyond the expected level
may indicate excessive compression.
Interpreting such changes requires a reference for how contact
is expected to evolve under the planned actions.
Tactile forecasting provides this reference, allowing incoming
measurements to be compared with earlier predictions during
execution~\cite{tacpac}.
Such verification assesses whether contact follows the current
plan, but leaves open whether that plan remains reliable under
the current visual conditions.

This motivates jointly considering visual reliability
and tactile forecast agreement (Fig.~\ref{fig:motivation}).
A plan perturbed by illumination may still produce the tactile
responses predicted for it, while unexpected contact can arise
even under reliable vision.
These complementary signals define four refinement
regimes.
When touch agrees with the forecast, low visual reliability
motivates \emph{visual compensation}, whereas high reliability
favors \emph{nominal preservation}.
When touch deviates from the forecast, low visual reliability
motivates stronger \emph{joint correction}, while high reliability
can still favor \emph{contact correction}.
Together, these regimes motivate us using visual reliability for refinement, with tactile forecast
discrepancies further increasing correction strength as contact
deviates from expectation. The question is \textbf{how to learn visual reliability and jointly refine actions.}

%combine it with tactile feedback to refine actions. 

%To translate these refinement regimes into execution-time action
%updates, 

In this paper, we present \textbf{TipDex-VTLA}, a predictive and reactive
VTLA framework that uses estimated visual reliability and tactile forecast
verification to gate tactile feature corrections.
Specifically, we introduce
\textbf{Visual Action Sensitivity Estimation (VASE)},
which learns a proxy for visual reliability from the action
discrepancies of a frozen VTLA policy across paired real normal and synthesized perturbed observations
that differ only in illumination.
For tactile-guided refinement, we develop
\textbf{Tactile Anticipation and Correction during Execution (TACE)},
which compares incoming tactile measurements with temporally
aligned forecasts consistent with the intervening executed actions.
The estimated visual reliability and verified tactile discrepancies jointly control
the strength of bounded feature corrections computed from current
tactile feedback and execution context.
By reusing cached flow states, the fast branch refines 
replaceable, unexecuted actions while preserving their original
execution times.

We evaluate TipDex-VTLA on four real-world dexterous manipulation
tasks and the UniVTAC simulation benchmark~\cite{chen2026univtac}.
Across the four real-world tasks, our TipDex-VTLA achieves average
full-task success rates of 87.5\% under standard lighting and
73.8\% under challenging unseen lighting.
These results demonstrate the effectiveness of reliability-aware
tactile refinement under illumination changes. The main contributions are as follows:
\begin{itemize}
    \item We develop TipDex-VTLA, a predictive
    and reactive VTLA framework that effectively refines actions under practical yet challenging unseen illumination variations.
 
    \item We introduce the concept of action-grounded
    visual reliability via VASE and jointly regulate tactile refinement using verified tactile forecast discrepancies via TACE.  

    \item Experiments on UniVTAC benchmark~\cite{chen2026univtac} and four real-world dexterous manipulation tasks demonstrate TipDex-VTLA's effectiveness and significantly improved performance
    under unseen illumination.
     % We evaluate \textbf{TipDex-VTLA} on four real-world
    % dexterous manipulation tasks and the UniVTAC
    % simulation benchmark~\cite{chen2026univtac},
    % demonstrating improved real-world full-task success
    % under unseen illumination.
\end{itemize}

%% file: sec/relatedwork.tex
\section{Related Work}
\label{sec:related_work}

\subsection{Vision-Language-Action Models}

Vision--language--action (VLA) models map visual observations
and language instructions to robot actions.
RT-1~\cite{rt1} scales multi-task learning from diverse robot
demonstrations, while RT-2~\cite{rt2} transfers pretrained
vision--language knowledge to control through joint fine-tuning
on robotic and web-scale vision--language data.
Open X-Embodiment~\cite{openx} aggregates demonstrations across
robots, tasks, and environments to support cross-embodiment
learning.
Octo~\cite{octo} develops a transformer-based diffusion policy
that supports adaptation to new observation and action spaces.
OpenVLA~\cite{openvla} adapts a pretrained vision--language model
for robotic control and enables parameter-efficient fine-tuning
for new robot setups.

For continuous action generation, $\pi_0$~\cite{pi0} combines
a pretrained vision--language backbone with a flow-matching
action expert to generate action chunks.
Building on this architecture, $\pi_{0.5}$~\cite{pi05} improves
open-world generalization through heterogeneous co-training
with robot demonstrations, web data, and high-level semantic
prediction.
These advances provide a foundation for incorporating tactile
observations, which complement visual scene understanding
with local contact cues.

\subsection{Vision-Tactile-Language-Action Models}

Vision--tactile--language--action (VTLA) models extend
generalist robot policies with physical feedback to
improve contact-rich manipulation.
Prior works~\cite{sparsh,seetotouch} explore tactile
representations and multimodal fusion across sensing modalities.
AnyTouch~\cite{anytouch} learns unified static-dynamic representations
across  visuo-tactile sensors.
ForceVLA~\cite{forcevla} and ForceVLA2~\cite{forcevla2} incorporate force-aware feedback
into robot action generation,
while DECO~\cite{deco} introduces a lightweight tactile adapter
for pretrained policies.
FTP-1~\cite{ftp1} learns transferable tactile
representations across sensors and embodiments.
Beyond direct fusion, AT-VLA~\cite{atvla} combines
adaptive tactile injection with dual-stream feedback
control, while TacVLA~\cite{tacvla} activates tactile
tokens according to contact states.
T-Rex~\cite{trex} enables high-frequency tactile
refinement through cached flow states and
variable-rate action generation.

Recent studies~\cite{vtam,vtwam,dextacwam} extend tactile feedback from current
observations to anticipated physical
interactions.
$\mathcal{N}_0$-VTLA~\cite{n0vtla} predicts latent tactile tokens
to condition action generation,
while ForeTac-VLA~\cite{foretacvla} investigates multi-step tactile
forecasting.
UniTacVLA~\cite{unitacvla} jointly models tactile understanding and
future contact prediction.
These approaches~\cite{tactilewam,tacdynwam} demonstrate that predicted tactile
responses can provide useful information before
physical contact is fully observed.

% More recent methods~\cite{hitacwam} combine tactile
% prediction with execution-time action refinement.
% TouchWorld~\cite{touchworld} integrates tactile
% prediction with high-frequency residual feedback,
% while ReTouch~\cite{retouch} refines tactile predictions
% and actions using incoming measurements.
% TacPAC~\cite{tacpac} compares observed touch with cached
% contact predictions to correct unexecuted actions.
% ResTacVLA~\cite{restacvla} represents touch as residuals
% relative to visual predictions and adaptively gates
% their integration.
% While these studies advance tactile prediction,
% adaptive gating, and reactive correction, they do not
% explicitly address reliability under unseen illumination variations.

Recent methods~\cite{hitacwam} combine tactile prediction
with execution-time action refinement.
TouchWorld~\cite{touchworld} couples tactile prediction
with high-frequency residual feedback, while
ReTouch~\cite{retouch} updates tactile predictions and
actions using incoming measurements.
TacPAC~\cite{tacpac} compares observed touch with cached
contact predictions to correct unexecuted actions, while
ResTacVLA~\cite{restacvla} adaptively gates tactile
residuals relative to visual predictions.
However, these methods do not explicitly address
reliability under unseen illumination changes.

\subsection{Visual Robustness in Robotic Manipulation}
Visual distribution shifts~\cite{liftfeat,traqpoint} remain a major challenge
for deploying robotic policies beyond their training
environments.
General robustness studies, including
ImageNet-C~\cite{imagenetc}, AugMix~\cite{augmix}, and DeepAugment~\cite{deepaugment}, investigate
model sensitivity to visual corruptions and
augmentation-based generalization.
In robotic manipulation, LIBERO-Plus~\cite{liberoplus} systematically
evaluates VLA policies under changes in illumination,
camera viewpoints, and other environmental factors.

Recent approaches improve VLA robustness through
different strategies. RobustVLA~\cite{robustvla} introduces robustness-aware
post-training, while StableVLA~\cite{stablevla} uses an information bottleneck
adapter to suppress irrelevant visual noise. ChromaGuard~\cite{chromaguard} studies illumination robustness while preserving task-relevant color
information. E-VLA~\cite{evla} incorporates
event-based sensing under degraded visual conditions.
Beyond visual feature robustness,
Evidence-Gated Regularization~\cite{egr} uses task-relevance
signals to regulate sensor dependence,
while BAS-VLA~\cite{basvla} investigates action calibration under
task-preserving appearance changes and task-semantic
variations.

Building on these studies, we combine action-aware
visual reliability assessment with tactile feedback
as complementary physical evidence.
This integration supports robust manipulation under
unseen illumination variations.

%% file: sec/method.tex
\begin{figure*}[!ht]
    \centering
    \includegraphics[width=\textwidth]   
   {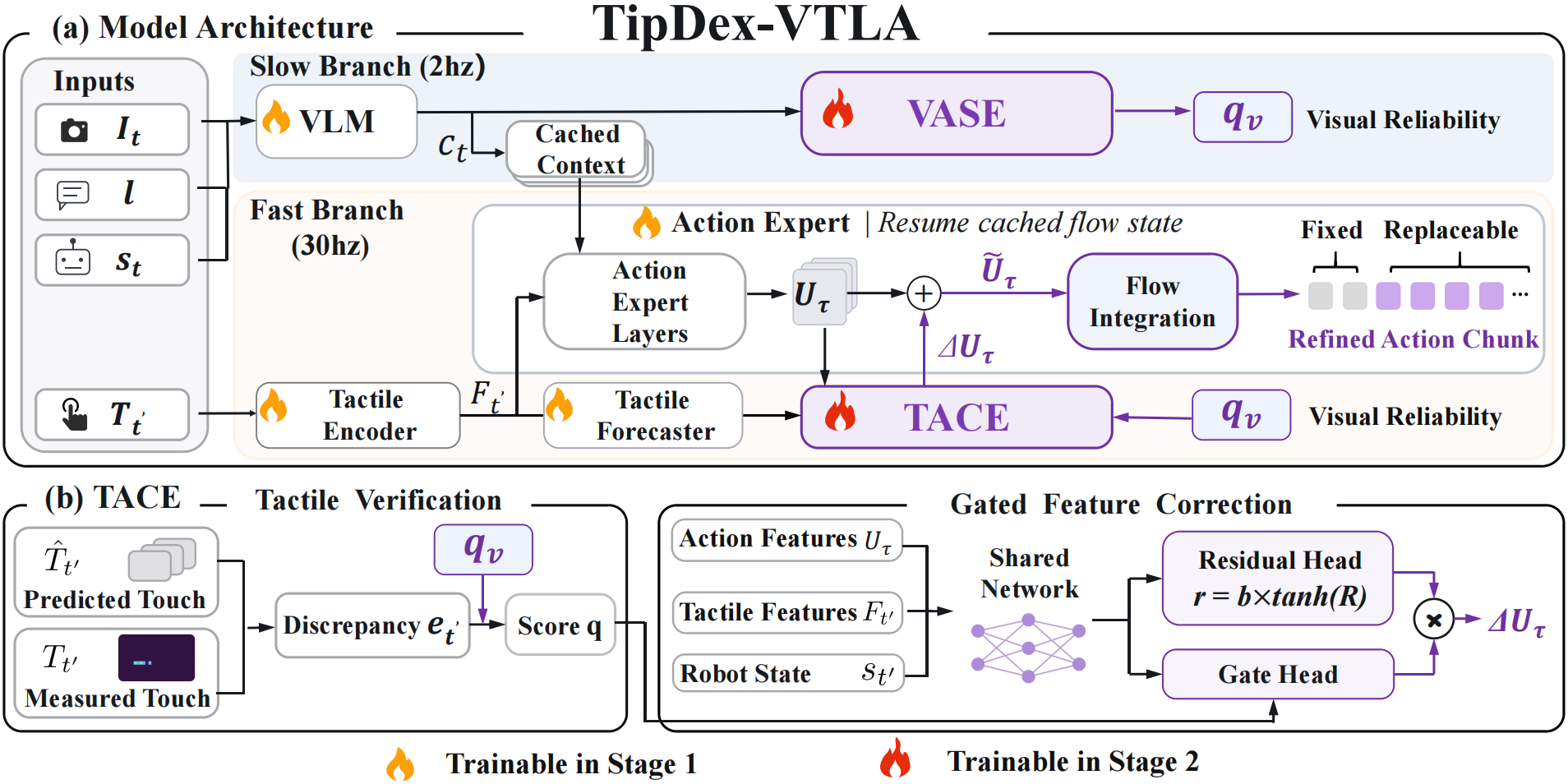}
    \caption{\textbf{Overview of TipDex-VTLA.}
 (a) The slow–fast architecture integrates a VLM, Tactile Encoder, and Action Expert with visual reliability estimation (VASE) and tactile forecasting. (b) TACE combines visual reliability and tactile forecast discrepancies to gate bounded corrections to Action Expert features.}
    \label{fig:architecture}
\end{figure*}

\section{Method}
\label{sec:method}

% Fig.~\ref{fig:architecture} summarizes the architecture and
% execution-time refinement of TipDex-VTLA.
% At observation time $t$, the base policy $\pi_\theta$ of
% TipDex-VTLA takes multi-view RGB images $I_t$, a language
% instruction $\ell$, robot
% proprioceptive state $s_t$, and recent tactile history
% $T_{\leq t}$ as inputs, where $T_t$ denotes a tactile frame.
% A Vision--Language Model (VLM) encodes the image--language--state
% prefix into $c_t$, while a Tactile Encoder produces observed
% tactile features $F_t$; these representations condition the
% base policy's Action Expert (Sec.~\ref{sec:architecture}).
% In parallel, Visual Action Sensitivity Estimation (VASE)
% reads $c_t$ to estimate visual reliability $q_v$
% (Sec.~\ref{sec:visual_usability}).
% Tactile Anticipation and Correction during Execution (TACE)
% combines this score with tactile forecast verification to
% regulate execution-time feature corrections
% (Sec.~\ref{sec:tactile_fusion}).
% The Action Expert generates a nominal chunk
% $\mathbf A_t$ of $H$ actions and uses these corrections
% to produce its updated version $\widetilde{\mathbf A}_t$.
% Both chunks are indexed by their original observation time $t$;
% fast updates preserve the original action positions and execution times.
% The model is trained in two stages, illustrated in
% Fig.~\ref{fig:training} and detailed in
% Sec.~\ref{sec:training_objectives}.

TipDex-VTLA integrates a multimodal base policy with
reliability-aware tactile refinement during execution
(Fig.~\ref{fig:architecture}).
At observation time $t$, the base policy $\pi_\theta$ receives
multi-view RGB images $I_t$, a language instruction $\ell$,
robot proprioceptive state $s_t$, and recent tactile history $T_{\leq t}$, where $T_t$ denotes a tactile frame.
The VLM encodes the image--language--state prefix into
context $c_t$, and the Tactile Encoder extracts observed
tactile features $F_t$.
Together, these representations condition the Action Expert
to generate a nominal chunk $\mathbf A_t$ of $H$ actions
(Sec.~\ref{sec:architecture}).
VASE estimates action-relevant visual reliability $q_v$
from $c_t$ (Sec.~\ref{sec:visual_usability}).
During execution, TACE verifies incoming tactile measurements
against temporally aligned forecasts and combines the resulting
discrepancies with $q_v$ to gate bounded feature corrections
(Sec.~\ref{sec:tactile_fusion}).
The fast branch reuses cached flow states to incorporate
these corrections into replaceable, unexecuted actions,
yielding an updated chunk $\widetilde{\mathbf A}_t$ while
preserving the original action positions and execution times.
Training proceeds in two stages, as illustrated in
Fig.~\ref{fig:training} and detailed in
Sec.~\ref{sec:training_objectives}.

\subsection{Model Architecture}
\label{sec:architecture}

The backbone of $\pi_\theta$ comprises a pretrained VLM,
a Tactile Encoder, and a flow-matching Action Expert~\cite{flowmatching}
(Fig.~\ref{fig:architecture}(a)).
The VLM processes the prefix formed by $(I_t,\ell,s_t)$
to produce $c_t$. The Tactile Encoder aggregates recent
tactile observations into $F_t$, preserving five fingertip
regions and three palm sensing regions.
The observed tactile features $F_t$ are concatenated with
$c_t$ to condition the Action Expert.
VASE branches from $c_t$ before this concatenation:
a lightweight VASE Head pools the prefix representation and
outputs $q_v$, which is cached for fast refinement.

The Action Expert adopts a slow--fast flow generation scheme
based on T-Rex~\cite{trex}.
We use $\tau$ for flow time, with noise at $\tau=0$
and actions at $\tau=1$.
The slow update caches $c_t$ and an intermediate flow state
$X_{\tau_s}$, where $0<\tau_s<1$.
At a subsequent fast update $t'>t$, generation resumes
from $\tau_s$ to $1$, reusing the cached prefix and
incorporating current tactile feedback.
Within this execution scheme, TACE applies reliability-aware
feature corrections to refine replaceable, unexecuted actions.
The original action positions and execution times remain fixed
throughout this process.
Implementation details of the slow--fast execution scheme
are provided in the supplementary material.

An auxiliary Tactile Forecaster $P$ shares the Tactile Encoder with
the Action Expert. Its inputs are $c_t$ and the current execution
context $\xi_{t'}$, which includes the robot state $s_{t'}$,
tactile features $F_{t'}$, and the
current action sequence with its execution times.
After each action update at $t'$, $P$ predicts the next $H$
normalized tactile frames, denoted by $\widehat{\mathcal T}_{t'}$
and cached for later verification against time-aligned
measurements, yielding a discrepancy signal for TACE.

TACE consists of a shared Refinement Network, a Residual Head,
and a Gate Head. It receives current tactile and execution
information, Action Expert features, $q_v$, and the verified
tactile discrepancy.
At each remaining flow step $\tau$, TACE modifies the
Action Expert feature tensor $U_\tau\in\mathbb R^{H\times d_U}$.
The corrected features pass through
the remaining Action Expert layers to generate updated actions.
Only replaceable, unexecuted commands are submitted as
replacements in $\widetilde{\mathbf A}_t$.

\subsection{Visual Action Sensitivity Estimation (VASE)}
\label{sec:visual_usability}

\begin{figure}[t]
    \centering
    \includegraphics[width=\linewidth]
   {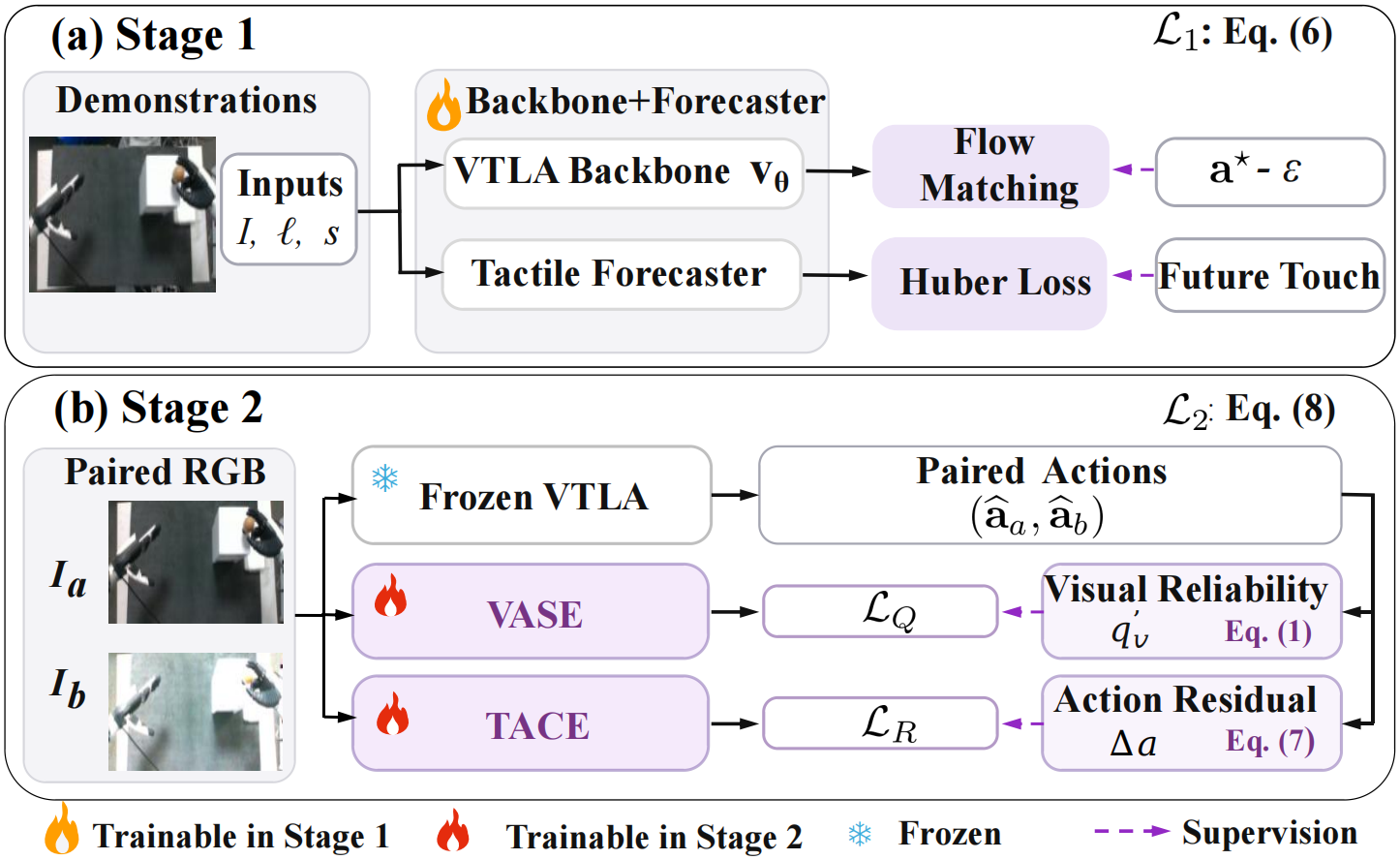}
    \caption{\textbf{Two-stage training of TipDex-VTLA.}
    (a) Demonstrations jointly train the VTLA backbone and
tactile forecaster.
    (b) Frozen-policy predictions under paired illumination
conditions supervise VASE reliability estimation and
TACE action refinement.}

    \label{fig:training}
\end{figure}

VASE estimates visual action reliability using the sensitivity
of a fixed reference policy to illumination changes.
For each demonstration observation at $t$, we retain the original
multi-view image $I_a$ and construct its paired variant $I_b$
by applying illumination augmentation to the same observation.
The paired-image sampling distribution assigns equal probability
($20\%$ each) to five conditions: low illumination, strong
illumination, changed illumination direction, uneven illumination,
and unchanged illumination. The pairing preserves the viewpoint
and physical state.
% AUTHOR CHECK: Specify the image-generation operators and their
% parameter ranges, and whether augmentation parameters are shared
% across views and time. The 20% sampling probability is confirmed.
We use lower-case action symbols for normalized chunks.
The frozen base policy obtained in Stage~1 generates
$\widehat{\mathbf a}_a$ and
$\widehat{\mathbf a}_b$ using identical language, robot state,
tactile history, sampling noise, and sampling settings.
Their difference therefore measures the reference policy's
response to the image perturbation under fixed non-image conditions.
As illustrated in Fig.~\ref{fig:training}(b), the same paired
predictions supply the supervision targets for both VASE and TACE.

We convert the paired action discrepancy into a supervision
target for $I_b$:
\begin{equation}
q_v^{\prime}
=
\exp\left(
-\frac{\operatorname{MSE}
\left(\widehat{\mathbf a}_a,\widehat{\mathbf a}_b\right)}
{\lambda}
\right).
\label{eq:usability_target}
\end{equation}
Here, $\operatorname{MSE}$ denotes mean squared error
between normalized action chunks, and $\lambda>0$
sets the decay scale.
The target $q_v^{\prime}\in(0,1]$ equals one when the paired
action predictions coincide and decreases as their
discrepancy grows.
The prime denotes a constructed supervision target.
It provides a proxy for visual reliability based on
the reference policy's action consistency under illumination
changes and does not represent action correctness or
task-success probability.

% The VASE Head $h_\omega$ maps the prefix representation
% to the visual reliability score:
% \begin{equation}
% q_v(I_t,\ell,s_t)
% =k+(1-k)\sigma\left(
% h_\omega\left(\operatorname{Pool}(c_t)\right)
% \right),
% \label{eq:usability_estimator}
% \end{equation}
% where $\sigma$ is the sigmoid function and $\omega$ denotes
% the head parameters. At deployment, this head uses the current
% prefix without a paired reference image and caches $q_v$
% at each slow update. Its training objective is given in Sec.~\ref{sec:training_objectives}.

The VASE Head $h_\omega$ maps the prefix representation
to a visual action reliability score:
\begin{equation}
q_v(I_t,\ell,s_t)
=
\sigma\left(
h_\omega\left(\operatorname{Pool}(c_t)\right)
\right),
\label{eq:usability_estimator}
\end{equation}
where $\sigma$ is the sigmoid function and $\omega$ denotes
the head parameters.
The score $q_v\in(0,1)$ estimates the frozen reference
policy's action consistency under illumination changes,
with larger values indicating greater predicted consistency.
It serves as a proxy for visual reliability.
At deployment, the head uses the current prefix without
requiring a paired reference image and caches $q_v$
at each slow update.
Its training objective is given in
Sec.~\ref{sec:training_objectives}.

\subsection{Tactile Anticipation and Correction during Execution (TACE)}
\label{sec:tactile_fusion}

As illustrated in Fig.~\ref{fig:architecture}(b), TACE uses
cached tactile predictions for verification.
The discrepancy and visual reliability gate
feature residuals computed from Action Expert features,
measured touch, and execution context.
The predictor path in Fig.~\ref{fig:architecture}(a)
summarizes this delayed verification.

\paragraph{Tactile Forecasting and Verification.}
At fast update $t'$, $\widehat T_{t'}$ denotes the latest
eligible cached prediction targeting $t'$, issued before $t'$
under commands consistent with the intervening executed actions.
We define $e_{t'}$ as the mean absolute error between
$\widehat T_{t'}$ and $T_{t'}$ after identical normalization.
If no eligible forecast exists, we set $e_{t'}=0$ to fall
back to the visual score; this does not indicate forecast accuracy.
After updating the action sequence, $P$ uses the updated
execution context to issue a new future tactile sequence:
\begin{equation}
\widehat{\mathcal T}_{t'}=P(c_t,\xi_{t'}).
\label{eq:tactile_forecast_verification}
\end{equation}
The sequence is cached for later verification.
Timestamp alignment, command-consistency checks, and
forecast invalidation are detailed in the supplementary material.
\paragraph{Gated Feature Correction.}
We combine the cached visual score and tactile discrepancy into
an execution score $q_{t'}=\mathcal G(q_v,e_{t'})\in(0,1)$.
The mapping $\mathcal G$ is non-decreasing in $q_v$ and
non-increasing in $e_{t'}$, with $\mathcal G(q_v,0)=q_v$.
Thus, visual reliability provides the baseline, while tactile
verification can lower the score. All action positions at a
fast update share this score.
% AUTHOR CHECK: Specify the implemented G (including numerical
% bounds and any parameters) in the implementation details.

For each replaceable, unexecuted action position $h$,
$R_\psi(U_{\tau,h},\xi_{t'})$ and
$S_\psi(U_{\tau,h},\xi_{t'})$ denote residual and gate-logit
predictions from the shared Refinement Network and their
respective heads. The Refinement Network uses only the state and tactile
components of $\xi_{t'}$; action history and the current
command schedule are used for tactile forecasting and verification. The correction is:
\begin{equation}
\begin{aligned}
g_{\tau,h}
&=\sigma\left(S_\psi(U_{\tau,h},\xi_{t'})
-\kappa\operatorname{logit}(q_{t'})\right),\\
\widetilde U_{\tau,h}
&=U_{\tau,h}
+g_{\tau,h}b\tanh\left(R_\psi(U_{\tau,h},\xi_{t'})\right).
\end{aligned}
\label{eq:tactile_correction}
\end{equation}
Here, $\psi$ denotes the refinement parameters, $b>0$
bounds each residual component, and $\kappa>0$ controls
the execution score's influence on the scalar gate
$g_{\tau,h}$.
For fixed network inputs, lower visual reliability or a larger
verified discrepancy favors a larger gate. A larger gate permits
stronger correction, while the actual feature change also depends
on the predicted residual. Other feature positions remain unchanged.

The remaining Action Expert layers convert $\widetilde U_\tau$
into refined actions. Only replaceable, unexecuted commands
are updated, preserving all other commands and the original
execution times. Paired-action supervision for the resulting
action change is introduced in Sec.~\ref{sec:training_objectives}.

\begin{figure*}[t!]
    \centering
    \scalebox{1}[0.8]{%
        \includegraphics[width=\textwidth]
        {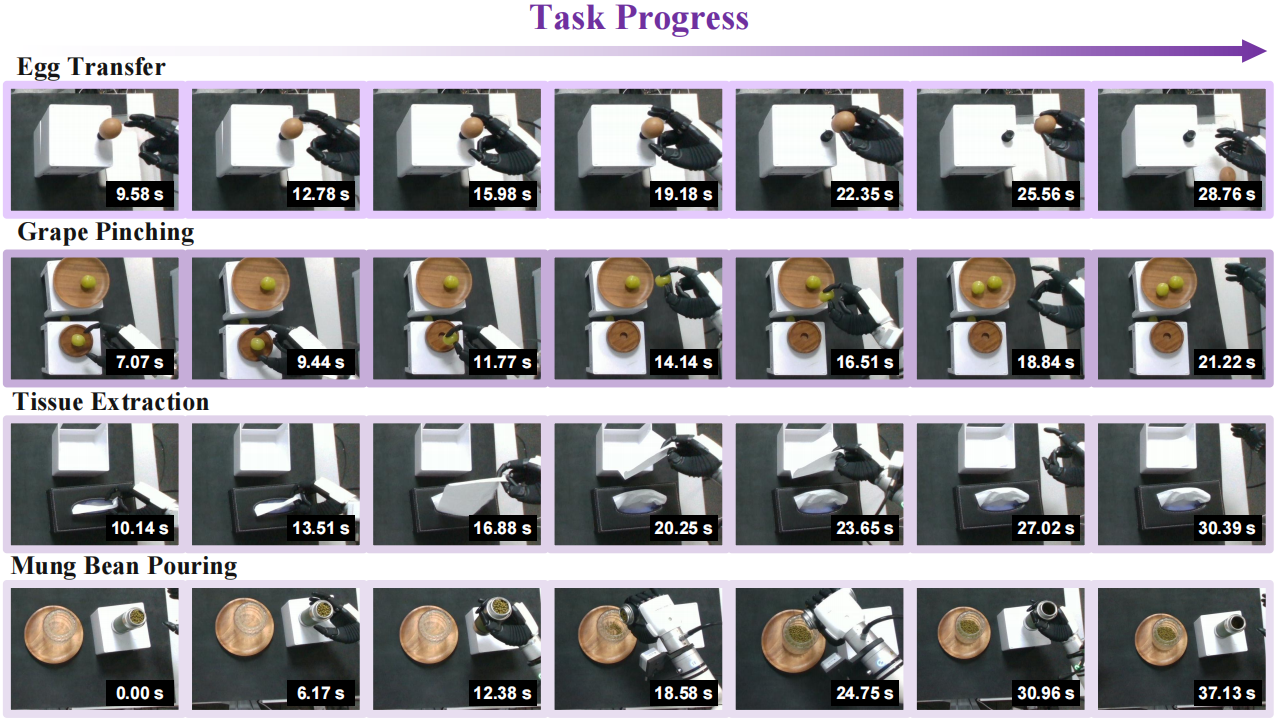}
    }
    \caption{\textbf{Real-world execution of TipDex-VTLA.}
    Temporally ordered keyframes illustrate four dexterous manipulation tasks:
    (a) egg transfer, (b) grape pinching, (c) tissue extraction,
    and (d) mung bean pouring.}
    \label{fig:real_world_execution}
\end{figure*}
\subsection{Two-Stage Training}
\label{sec:training_objectives}

Stage~1 establishes a tactile-conditioned base policy and
an auxiliary Tactile Forecaster. Stage~2 keeps these modules
fixed and trains VASE and TACE from paired illumination
observations. The same frozen reference predictions supply
the reliability targets and the action-correction targets.

\paragraph{Stage 1: VTLA Backbone and Tactile Forecasting.}
As shown in Fig.~\ref{fig:training}(a), we jointly train
the VTLA backbone and tactile forecaster using demonstrations
with unaugmented images. Expert actions supervise action
generation, while recorded future tactile measurements
supervise forecasting.
For a normalized expert action chunk $\mathbf a^\star$,
we sample $\epsilon\sim\mathcal N(0,\mathbf I)$ and
$\tau\sim\mathcal U(0,1)$ and define the interpolated
action state:
\begin{equation}
X_\tau=(1-\tau)\epsilon+\tau\mathbf a^\star.
\end{equation}
The joint objective is:
\begin{equation}
\begin{aligned}
\mathcal L_1=\mathbb E\bigl[
&\left\|
v_\theta(X_\tau,\tau\mid c_t,F_t)
-(\mathbf a^\star-\epsilon)
\right\|_2^2
+\lambda_T\mathcal L_T
\bigr].
\end{aligned}
\label{eq:stage1_objective}
\end{equation}
The expectation is over demonstrations, noise, and flow time;
the squared norm sums over action entries.
Here, \(\mathcal L_T\) is the Huber loss between predicted
and measured normalized tactile sequences, averaged over
future frames and tactile values, and \(\lambda_T>0\)
balances the two losses.
Forecast targets are paired with their recorded command
schedules. All Stage~1 parameters are then frozen.

\begin{table*}[t!]
    \centering
    \footnotesize
    \setlength{\tabcolsep}{2.5pt}       % 原来是 3pt
    \renewcommand{\arraystretch}{0.95} % 原来是 1.0

    \begin{tabularx}{0.90\textwidth}{   % 原来是 0.95\textwidth
        ll
        *{8}{>{\centering\arraybackslash}X}
        !{{\color{black!15}\vrule width 0.3pt}}
        >{\centering\arraybackslash}X
    }
        \toprule
        & & \multicolumn{8}{c}{\textbf{Task Success Rate (\%)}}
        & \textbf{Avg.} \\
        \cmidrule(lr){3-10}

        & \textbf{Method}
        & \textbf{\shortstack{Lift\\Bottle}}
        & \textbf{\shortstack{Pull-out\\Key}}
        & \textbf{\shortstack{Lift\\Can}}
        & \textbf{\shortstack{Put\\Bottle}}
        & \textbf{\shortstack{Insert\\Hole}}
        & \textbf{\shortstack{Insert\\HDMI}}
        & \textbf{\shortstack{Insert\\Tube}}
        & \textbf{\shortstack{Grasp\\Classify}}
        & \\
        \midrule

        \multirow{3}{*}{\textit{Without Tactile}}
        & ACT (Vision Only)~\cite{act}
        & 42.0 & 28.0 & 20.0 & 28.0
        & 19.0 & 15.0 & 45.0 & 50.0
        & 30.9 \\

        & $\pi_{0.5}$~\cite{pi05}
        & \textbf{100.0} & 35.0 & 6.0 & 34.0
        & 25.0 & 8.0 & 74.0 & 49.0
        & 41.4 \\

        & LingBot-VA~\cite{lingbotva2026}
        & 0.0 & 58.0 & 0.0 & 0.0
        & 42.0 & \textbf{38.0} & \underline{96.0} & 17.0
        & 31.4 \\

        \midrule

        \multirow{6}{*}{\textit{With Tactile}}
        & ACT + UniVTAC~\cite{chen2026univtac}
        & 71.0 & 46.0 & 29.0 & 31.0
        & 24.0 & 28.0 & 56.0 & \underline{99.0}
        & 48.0 \\

        & VITaL~\cite{vital}
        & 72.0 & 47.0 & 8.0 & 32.0
        & 25.0 & 6.0 & 34.0 & \textbf{100.0}
        & 40.5 \\

        & FTP-1~\cite{ftp1}
        & 97.0 & 48.0 & 65.0 & 47.0
        & 64.0 & 18.0 & 79.0 & \textbf{100.0}
        & 64.8 \\

        & T-Rex~\cite{trex}
        & \underline{99.0} & \textbf{99.0} & 75.0 & \underline{79.0}
        & \textbf{100.0} & 8.0 & 95.0 & 95.0
        & 81.3 \\

        & $\mathcal{N}_0$-VTLA~\cite{n0vtla}
        & \underline{99.0} & \textbf{99.0} & \textbf{88.0} & 60.0
        & \underline{95.0} & 25.0 & \textbf{99.0}
        & \textbf{100.0}
        & \underline{83.1} \\

        \cmidrule(lr){2-11}
        \rowcolor{AvgTint}
        \cellcolor{white}
        & \textbf{TipDex-VTLA}
        & \textbf{100.0} & \underline{97.0} & \underline{87.0}
        & \textbf{80.0} & 89.0 & \underline{30.0}
        & \underline{96.0} & \textbf{100.0}
        & \textbf{84.9} \\

        \bottomrule
    \end{tabularx}

    \caption{
    Full-task success rates (\%) on the eight UniVTAC
    manipulation tasks, grouped by tactile input availability.
    Published baselines retain their originally reported
    configurations and evaluation results.
    Bold denotes column-wise best results.
    }
    \label{tab:univtac_results}
\end{table*}

\paragraph{Stage 2: Reliability-Aware Refinement.}
As shown in Fig.~\ref{fig:training}(b), we optimize the VASE
Head \(\omega\) and TACE modules \(\psi\) while keeping all
Stage~1 parameters fixed.
VASE uses the paired illumination observations described
in Sec.~\ref{sec:visual_usability}; TACE additionally uses
their recorded tactile and execution histories.
The target \(q_v^{\mathrm{'}}\) from
Eq.~\eqref{eq:usability_target} supervises the VASE score for \(I_b\).
The same frozen reference predictions provide TACE's
signed action-correction target:
\begin{equation}
\Delta\mathbf a
=\widehat{\mathbf a}_a-\widehat{\mathbf a}_b.
\label{eq:paired_residual_target}
\end{equation}
Using \(I_b\), we cache $X{\tau_s}$ from the same nominal
trajectory that produces \(\widehat{\mathbf a}_b\).
TACE resumes generation from \(\tau_s\) to \(1\) with recorded
feedback, yielding the normalized refined action chunk
\(\widetilde{\mathbf a}_b\).
Forecast verification and execution scoring follow
Sec.~\ref{sec:tactile_fusion}.
All action outputs use the same normalization and remain
aligned to their original execution times.

Let \(M\) select replaceable, unexecuted action coordinates.
The two losses and their joint objective are:
\begin{equation}
\begin{aligned}
\mathcal L_Q
&=\mathbb E\left[
\operatorname{BCE}\left(q_v(I_b,\ell,s_t),q_v^{\prime}\right)
\right],\\
\mathcal L_R
&=\mathbb E\left[
\operatorname{MSE}_M\left(
\widetilde{\mathbf a}_b-\widehat{\mathbf a}_b,\Delta\mathbf a
\right)\right],\\
\mathcal L_2
&=\lambda_Q\mathcal L_Q+\lambda_R\mathcal L_R.
\end{aligned}
\label{eq:stage2_objective}
\end{equation}

Expectations are over training samples.
\(\operatorname{BCE}\) denotes binary cross-entropy, and
\(\operatorname{MSE}_M\) averages over action coordinates
selected by \(M\).
Samples with no selected coordinates are excluded from
\(\mathcal L_R\).
The paired target \(\Delta\mathbf a\) is fixed, and
\(\lambda_Q,\lambda_R>0\) balance the two losses.

During refinement training, the gate uses
\(\operatorname{sg}(q_{t'})\), where \(\operatorname{sg}\)
stops gradients.
Thus, \(\mathcal L_Q\) updates only \(\omega\), while
\(\mathcal L_R\) supervises the action change induced by
Eq.~\eqref{eq:tactile_correction} and backpropagates through
the remaining generation steps to update only \(\psi\),
keeping the backbone frozen.

%% file: sec/experiments_fix.tex
\section{Experiments}
\label{sec:experiments}

We evaluate TipDex-VTLA on four real-world dexterous manipulation tasks
and in UniVTAC simulation~\cite{chen2026univtac}. We assess robustness
to unseen lighting and analyze the cues that govern execution-time
tactile refinement.

\subsection{Experimental Setup}
\label{sec:exp_setup}

\paragraph{Platforms and Data.}
Our platform consists of two RealMan RM75 7-DoF arms equipped with
Inspire RH56F2-E2-T1 dexterous hands. Each hand has eight piezoresistive tactile arrays
comprising 364 taxels, sampled at 30 Hz. We capture multi-view RGB
observations using one overhead RealSense D405 and two wrist-mounted
D435i cameras.

We use 300 demonstration trajectories for training:
200 to train the base policy and tactile forecaster in Stage~1,
and 100 additional trajectories to train VASE and TACE in Stage~2.
Each Stage~2 trajectory is paired with its illumination-perturbed
counterpart.
Detailed data construction procedures and illumination
perturbation settings are provided in the supplementary
material.
\begin{table}[t]
    \centering
    \footnotesize
    \setlength{\tabcolsep}{3pt}
    \renewcommand{\arraystretch}{1.0}

    \begin{tabularx}{0.90\linewidth}{
        l
        *{4}{>{\centering\arraybackslash}X}
        !{{\color{black!15}\vrule width 0.25pt}}
        >{\centering\arraybackslash}X
    }
        \toprule
        \textbf{Method}
        & \textbf{Egg}
        & \textbf{Grape}
        & \textbf{Tissue}
        & \textbf{Beans}
        & \textbf{Avg.} \\
        \midrule

        \multicolumn{6}{l}{
            \textit{Standard lighting (N)}} \\
        \addlinespace[2pt]

        ACT~\cite{act}
        & 0.0 & 30.0 & 20.0 & 25.0 & 18.8 \\

        $\pi_{0.5}$~\cite{pi05}
        & 20.0 & 80.0 & 40.0 & \underline{80.0} & 55.0 \\

        % AT-VLA~\cite{atvla}
        % & 60.0 & 40.0 & 65.0 & 60.0 & 56.3 \\

        T-Rex~\cite{trex}
        & \underline{65.0} & 80.0 & 60.0 & \underline{80.0} & \underline{71.3} \\

        $\mathcal{N}_0$-VTLA~\cite{n0vtla}
        & 60.0 & \underline{85.0} & \underline{70.0} & 60.0 & 68.8 \\

        \rowcolor{AvgTint}
        \textbf{TipDex-VTLA}
        & \textbf{90.0} & \textbf{95.0} & \textbf{75.0}
        & \textbf{90.0} & \textbf{87.5} \\

        \midrule
        \multicolumn{6}{l}{
            \textit{Unseen lighting (U)}} \\
        \addlinespace[2pt]

        ACT~\cite{act}
        & 0.0 & 0.0 & 0.0 & 0.0 & 0.0 \\

        $\pi_{0.5}$~\cite{pi05}
        & 20.0 & \underline{30.0} & 25.0 & 20.0 & 23.8 \\

        % AT-VLA~\cite{atvla}
        % & 10.0 & 25.0 & 30.0 & 15.0 & 20.0 \\

        T-Rex~\cite{trex}
        & 20.0 & \underline{30.0} & 30.0 & 10.0 & 22.5 \\

        $\mathcal{N}_0$-VTLA~\cite{n0vtla}
        & \underline{30.0} & 20.0 & \underline{35.0} & \underline{40.0} & \underline{31.3} \\

        \rowcolor{AvgTint}
        \textbf{TipDex-VTLA}
        & \textbf{70.0} & \textbf{90.0} & \textbf{60.0}
        & \textbf{75.0} & \textbf{73.8} \\

        \bottomrule
    \end{tabularx}

    \caption{
    Real-world full-task success rates (\%) across four
    dexterous manipulation tasks under standard lighting (N)
    and unseen lighting conditions (U).
    }
    \label{tab:real_results}
\end{table}
\paragraph{Training and Implementation.}
In Stage~1, we jointly train the VTLA backbone, tactile encoder,
and tactile forecaster for 20,000 optimization steps. In Stage~2,
we freeze these modules and optimize only VASE and TACE for
10,000 steps. Both stages use eight NVIDIA A100 GPUs and the
AdamW optimizer, with a global batch size of 128 and a peak
learning rate of $5\times10^{-5}$.
We use ten uniform flow integration steps.
Fast tactile refinement is applied during the final four
steps, from $\tau_s=0.6$ to $\tau=1$.
We bound each feature residual component by $b = 0.1$.
All model parameters remain fixed during deployment.

% \paragraph{Platforms and data.}
% Our real-world platform consists of two RealMan RM75
% 7-DoF robotic arms equipped with Inspire RH56F2-E2-T1
% dexterous hands operating in joint-position control.
% The perception system comprises one top-mounted
% RealSense D405 camera and two wrist-mounted
% RealSense D435i cameras, providing multi-view RGB
% observations. Piezoresistive tactile arrays are
% sampled at 30 Hz.
% We collect 200 demonstrations per task, divided
% into two disjoint sets of 100 for Stage~1 and Stage~2,
% with validation and test trajectories held out.

% \paragraph{Training and implementation.}

% Following Sec.~\ref{sec:training_objectives},
% we train the VTLA backbone and tactile forecaster
% for 20,000 steps, then freeze them and optimize
% VASE and TACE refinement for 10,000 steps.
% Training uses eight NVIDIA A100 GPUs, AdamW,
% a batch size of 128, and a peak learning rate
% of $5\times10^{-5}$ in both stages.
% We use an action horizon of $H=30$ and ten
% uniform flow integration steps, with the final
% four steps allocated to fast tactile refinement
% ($\tau_s=0.4$).
% The feature-residual bound is $b_U=0.1$.In Stage~2, half of the demonstrations undergo
% RGB photometric augmentation while retaining
% their original action and tactile targets.
% All parameters remain fixed during deployment.
% Further implementation details are provided
% in the supplementary material.

\paragraph{Evaluation Protocol.}
For real-world evaluation, we adapt vision--tactile baselines
to our piezoresistive tactile sensors by integrating our tactile encoder. For each
method and task, we run 20 trials under standard indoor lighting (N)
and 20 under unseen lighting (U). The 20 unseen configurations span
ten combinations of asymmetric lighting and color temperature,
five low-light settings, and five bright-light settings. We match
task layouts and reset conditions and use automatic camera exposure. Further details of the experimental settings and evaluation
procedures are provided in the supplementary material.
We report full-task success rates (\%)  across tasks.

For simulation, we integrate the optical tactile encoder released
with UniVTAC into our VTLA backbone and evaluate under standard
simulated lighting. We use 100 test rollouts per policy per task. 
% \begin{table}[H]
%     \centering
%     \small
%     \begin{tabular}{lc}
%         \toprule
%         Training Stage & Demonstrations \\
%         \midrule
%         Stage~1 & 200 \\
%         \midrule
%         Stage~2: Standard lighting & 100 \\
%         Low illumination & 20 \\
%         Strong illumination & 20 \\
%         Changed illumination direction & 20 \\
%        Uneven illumination & 20 \\
%        Unchanged light & 20 \\
%         \midrule
%         Total & 400 \\
%         \bottomrule
%     \end{tabular}
%     \caption{
%     Distribution of real-world training demonstrations
%     per manipulation task across the two training stages.
%     }
%     \label{tab:data_distribution}
% \end{table}

\paragraph{Baselines.}
For real-world evaluation, we compare TipDex-VTLA with vision-only ACT~\cite{act} and $\pi_{0.5}$~\cite{pi05}, as well as tactile-enabled T-Rex~\cite{trex} and $\mathcal{N}_{0}$-VTLA~\cite{n0vtla}. For a fair comparison, all methods are trained on the complete dataset described above. For simulation, we compare with these four methods on all eight UniVTAC tasks~\cite{chen2026univtac}. We additionally include published results for vision-only LingBot-VA~\cite{lingbotva2026} and tactile-enabled VITaL~\cite{vital}, FTP-1~\cite{ftp1}, and ACT equipped with the UniVTAC tactile encoder~\cite{chen2026univtac}, retaining their reported configurations. The two ACT variants provide references with and without tactile conditioning within the same policy framework.

\begin{figure*}[t!]
    \centering
    \scalebox{1}[0.90]{%
        \includegraphics[width=\textwidth]
        {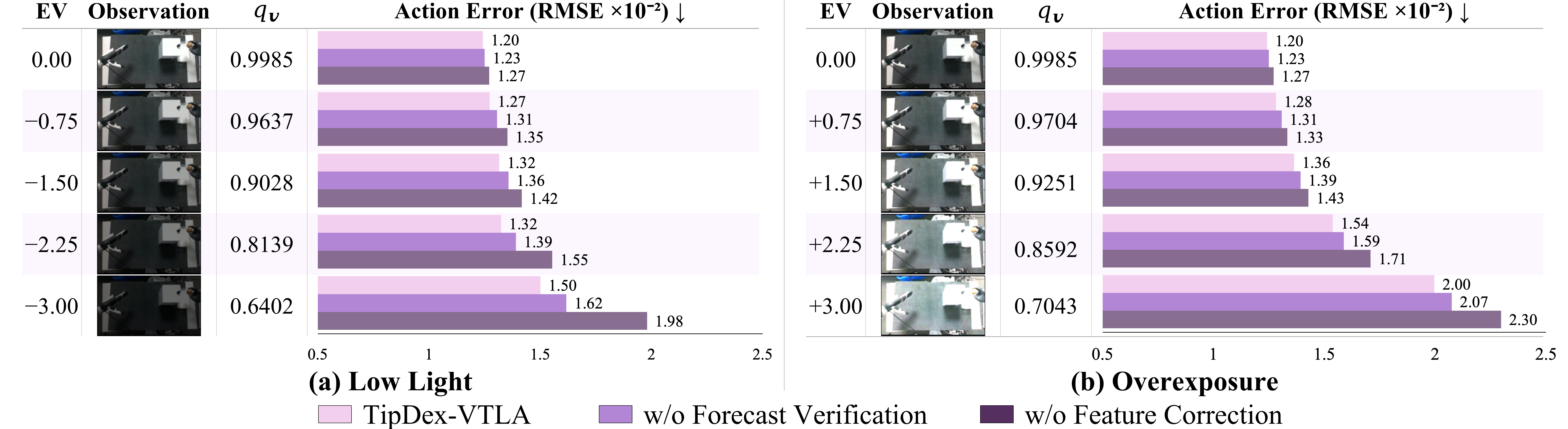}
    }
\caption{
\textbf{Visual reliability and trajectory prediction under
illumination changes.}
Visual reliability $q_v$ and RMSE between predicted and
ground-truth (GT) trajectories under (a) low light and
(b) overexposure, comparing TipDex-VTLA with variants
without forecast verification or feature correction. EV (exposure value) indicates exposure level; values represent EV offsets from the reference (0 EV).
}
    \label{fig:illumination_refinement}
\end{figure*}

% \begin{figure*}[t]
%     \centering
%     \includegraphics[width=\textwidth]
% {fig/ablation_new.png}

% \caption{\textbf{Visual reliability and action refinement under illumination changes.}
% Top: the original observation (Ori.) and perturbed examples.
% (a) VASE generally predicts lower visual reliability $q_v$ as perturbations intensify.
% (b) $\Delta\mathrm{RMSE}$ measures how much action refinement reduces the mean error relative to expert actions; larger positive values indicate greater improvement.}

%     \label{fig:visual_correction}
% \end{figure*}

\subsection{Real-World Results}
\label{sec:real_evaluation}

We evaluate four real-world tasks (Egg Transfer, Grape Pinching,
Tissue Extraction, and Mung Bean Pouring), involving fragile, soft,
deformable, and granular objects, respectively. 
Fig.~\ref{fig:real_world_execution} shows representative executions
under standard lighting. TipDex-VTLA achieves the highest success rate on all four tasks under
both lighting conditions, averaging 87.5\% under standard lighting and
73.8\% under unseen lighting (Table~\ref{tab:real_results}). Under unseen
lighting, it exceeds $\mathcal{N}_0$-VTLA by
42.5 percentage points.

\subsection{Simulation Results}
\label{sec:simulation_evaluation}

We evaluate TipDex-VTLA on all eight UniVTAC tasks with
both VASE and TACE enabled.
Optical tactile images are encoded by a shared
ResNet-18~\cite{resnet}, initialized from the released
UniVTAC tactile encoder~\cite{chen2026univtac}.
The tactile forecaster predicts future tactile images
for TACE verification.

As shown in Tab.~\ref{tab:univtac_results}, the full
adapted model achieves a mean success rate of 84.9\%. TACE compares forecasts with time-aligned measurements and combines discrepancies with VASE reliability to gate
Action Expert feature corrections for replaceable, unexecuted actions.

\begin{table}[t]
    \centering
    \footnotesize
    \setlength{\tabcolsep}{3pt}
    \renewcommand{\arraystretch}{1.0}

    \begin{tabularx}{0.9\linewidth}{
        l
        *{4}{>{\centering\arraybackslash}X}
        !{{\color{black!15}\vrule width 0.25pt}}
        >{\centering\arraybackslash}X
    }
        \toprule
        \textbf{Configuration}
        & \textbf{Egg}
        & \textbf{Grape}
        & \textbf{Tissue}
        & \textbf{Beans}
        & \textbf{Avg.} \\
        \midrule

        \multicolumn{6}{l}{
            \textit{Standard lighting (N)}} \\
        \addlinespace[2pt]

        \rowcolor{AvgTint}
        \textbf{Full TipDex-VTLA}
        & \textbf{90.0}
        & \textbf{95.0}
        & \textbf{75.0}
        & \textbf{90.0}
        & \textbf{87.5} \\

        w/o VASE
        & 85.0 & 95.0 & 70.0 & 90.0 & 85.0 \\

        w/o Tactile Verification
        & 80.0 & 85.0 & 60.0 & 90.0 & 78.8 \\

        w/o Gate Head
        & 80.0 & 85.0 & 65.0 & 85.0 & 78.8 \\

        \midrule
        \multicolumn{6}{l}{
            \textit{Unseen lighting (U)}} \\
        \addlinespace[2pt]

        \rowcolor{AvgTint}
        \textbf{Full TipDex-VTLA}
        & \textbf{70.0}
        & \textbf{90.0}
        & \textbf{60.0}
        & \textbf{75.0}
        & \textbf{73.8} \\

        w/o VASE
        & 45.0 & 40.0 & 55.0 & 30.0 & 42.5 \\

        w/o Tactile Verification
        & 65.0 & 80.0 & 45.0 & 60.0 & 62.5 \\

        w/o Gate Head
        & 30.0 & 45.0 & 40.0 & 45.0 & 40.0 \\

        \bottomrule
    \end{tabularx}

    \caption{
    Component ablation under standard (N) and unseen (U)
    real-world lighting.
    Values represent full-task success rates (\%).
    }
    \label{tab:ablation}
\end{table}

\subsection{Ablation Studies}
\label{sec:ablation}

% \paragraph{Component Ablation.}
% Tab.~\ref{tab:ablation} compares components under both lighting
% conditions. Under standard lighting, the full model achieves
% 87.5\% average success, compared with 85.0\% without VASE
% and 78.8\% when removing either tactile verification or the
% Gate Head.
% Under unseen lighting, removing VASE, tactile verification,
% or the Gate Head yields 42.5\%, 62.5\%, and 40.0\%,
% respectively, compared with 73.8\% for the full model.
% The larger performance drop without VASE under unseen lighting
% supports its role in illumination robustness, while tactile
% verification and gating also benefit standard-light performance.
% Combining both reliability cues yields an 11.3-percentage-point
% gain over the variant without tactile verification under
% unseen lighting.

\paragraph{Component Ablation.}
Tab.~\ref{tab:ablation} compares components under both lighting
conditions. Under standard lighting, the full model achieves
87.5\% average success, compared with 85.0\% without VASE
and 78.8\% without either tactile verification or the
Gate Head.
Under unseen lighting, removing VASE, tactile verification,
or the Gate Head yields 42.5\%, 62.5\%, and 40.0\%,
respectively, compared with 73.8\% for the full model.
The larger performance drop without VASE under unseen lighting
supports its role in illumination robustness, while tactile
verification and gating benefit standard-light performance.
Combining both reliability cues yields an 11.3-point
gain over the variant without tactile verification under
unseen lighting.

\paragraph{Action refinement under illumination changes.}
Fig.~\ref{fig:illumination_refinement} shows decreasing $q_v$
and larger refinement gains as illumination perturbations
intensify.
At $-3$ and $+3$ EV, the full model reduces trajectory RMSE
by $24.2\%$ and $13.0\%$ relative to disabling feature
correction; retaining forecast verification further reduces
error.
These results support the complementary cues motivating
the four regimes in Fig.~\ref{fig:motivation}.
With forecast agreement, reliable vision favors
\emph{nominal preservation}, while low visual reliability
motivates \emph{visual compensation}.
Forecast disagreement instead motivates \emph{contact
correction} with reliable vision and \emph{joint correction}
with low visual reliability.

%% file: sec/conclusion.tex
\section{Conclusion}
We proposed TipDex-VTLA, a  policy that addresses
action drift induced by changes in the visual environment
during dexterous manipulation.
The framework comprises two modules: VASE and TACE.
VASE estimates visual action reliability from policy sensitivity
to illumination changes.
TACE combines this estimate with tactile prediction discrepancies
to regulate execution-time refinement of unexecuted actions.
Evaluation on four real-world tasks and the UniVTAC benchmark
supports the effectiveness of this approach, with improved
real-world full-task success under unseen lighting.
The current reliability estimate remains a policy-dependent
proxy learned from illumination perturbations.
Future work will explore broader visual disturbances and
task-level feedback to improve reliability estimation and
extend tactile-guided refinement to more diverse manipulation
scenarios.